**RESEARCH**

# Enhanced prediction of spine surgery outcomes using advanced machine learning techniques and oversampling methods

José Alberto Benítez-Andrades[1,2]*[†] , Camino Prada-García[3,4†], Nicolás Ordás-Reyes[5†], Marta Esteban Blanco[6†], Alicia Merayo[5†] and Antonio Serrano-García[2,7†]

**Abstract**

**Purpose:** Accurate prediction of spine surgery outcomes is essential for optimizing treatment strategies. This study presents an enhanced machine learning approach to classify and predict the success of spine surgeries, incorporating advanced oversampling techniques and grid search optimization to improve model performance.

**Methods:** Various machine learning models, including GaussianNB, ComplementNB, KNN, Decision Tree, KNN with RandomOverSampler, KNN with SMOTE, and grid-searched optimized versions of KNN and Decision Tree, were applied to a dataset of 244 spine surgery patients. The dataset, comprising pre-surgical, psychometric, socioeconomic, and analytical variables, was analyzed to determine the most efficient predictive model. The study explored the impact of different variable groupings and oversampling techniques.

**Results:** Experimental results indicate that the KNN model, especially when enhanced with RandomOverSampler and SMOTE, demonstrated superior performance, achieving accuracy values as high as 76% and an F1-score of 67%. Grid-searched optimized versions of KNN and Decision Tree also yielded significant improvements in predictive accuracy and F1-score.

**Conclusions:** The study highlights the potential of advanced machine learning techniques and oversampling methods in predicting spine surgery outcomes. The results underscore the importance of careful variable selection and model optimization to achieve optimal performance. This system holds promise as a tool to assist healthcare professionals in decision-making, thereby enhancing spine surgery outcomes. Future research should focus on further refining these models and exploring their application across larger datasets and diverse clinical settings.

## Introduction and related work

Spine surgery is a critical intervention in the treatment of various spinal conditions, and accurate prediction of surgical outcomes is essential for optimizing treatment strategies. The variability in outcomes can be attributed not only to clinical and anatomical factors but also to a wide range of socio-economic and psychometric variables, which influence the recovery and satisfaction of patients. These factors include employment status, mental health conditions, and socio-economic background, which have been shown to significantly affect the success of spine surgeries by impacting post-operative recovery and long-term patient satisfaction [1–3].

Traditional methods for predicting spine surgery outcomes rely primarily on clinical evaluations, imaging studies, and patient-reported outcomes. However,

†José Alberto Benítez-Andrades, Camino Prada-García, Nicolás Ordás-Reyes, Marta Esteban Blanco, Alicia Merayo, Antonio Serrano-García have contributed equally to this work.

*Correspondence:  jbena@unileon.es
² Instituto de Investigación Biosanitaria de León (IBIOLEÓN), Calle Altos de Nava, s/n, 24008 León, Spain
Full list of author information is available at the end of the article these approaches are often subjective and may fail to accurately capture the complexity of surgical success. Recent advancements in machine learning provide a compelling alternative by leveraging large datasets and sophisticated algorithms to improve predictive accuracy. These models enable the integration of a broader range of variables, including socio-economic and psychometric data, which are frequently overlooked in traditional clinical assessments. For instance, factors such as psychological stress, depression levels, and financial stability can significantly impact post-operative recovery and overall patient satisfaction; when incorporated into machine learning models, these variables enhance the precision of outcome predictions [4].

Given the highly invasive nature of spine surgery, accurate outcome predictions are crucial not only for surgical planning but also for effective patient counseling and expectation management. By including socio-economic and psychometric variables, healthcare providers can develop tailored post-operative care plans that address the specific needs of each patient. For example, insights into a patient's employment status or psychological resilience can help anticipate their ability to return to work or their mental health trajectory, thus enabling more targeted interventions [5]. Moreover, accurately identifying high-risk patients early on can optimize resource allocation and reduce healthcare costs by focusing attention where surgery will likely have the greatest benefit.

The multifactorial nature of spine surgery outcomes-encompassing biological, psychological, and social factors-adds further complexity to the prediction process. Socio-economic and psychometric factors, such as mental health status, social support, and financial stability, intersect with biological variables to shape surgical success. Machine learning models that integrate these diverse parameters can provide a more holistic understanding of recovery determinants and long-term outcomes, surpassing the capabilities of traditional clinical predictors [1]. In this regard, recent research has demonstrated the effectiveness of such models by incorporating a wide range of clinical, psychometric, and socio-economic variables, underscoring the importance of factors like depression, anxiety, social support, income level, and healthcare access for achieving comprehensive predictive insights [6].

This approach not only improves predictive accuracy but also yields essential information for personalized care. By integrating psychometric assessments (e.g., depression and anxiety scores) with socio-economic data (e.g., job stability and education level), machine learning algorithms can generate more precise and nuanced predictions. Consequently, healthcare providers can design individualized rehabilitation plans and ultimately enhance clinical outcomes [7].

Recent advancements in machine learning (ML) have demonstrated their transformative potential in predicting outcomes across various medical domains, including cardiology, oncology, and orthopedics. ML models have been increasingly utilized to analyze complex, high-dimensional datasets to predict disease progression, treatment response, and surgical outcomes. For instance, convolutional neural networks (CNNs) have been applied to imaging data to predict cardiovascular events with high accuracy [8], while ensemble models, such as random forests and gradient boosting, have shown promise in stratifying cancer patients based on survival probabilities [9]. In orthopedic surgery, predictive models like support vector machines and deep learning architectures have been developed to estimate postoperative complications and recovery trajectories [10]. These applications highlight the ability of ML to integrate diverse data sources, including clinical, genetic, and imaging data, offering a comprehensive approach to personalized medicine.

In the context of spine surgery, ML adoption is still in its early stages but has begun to show significant promise. Several studies have explored the use of ML models to predict outcomes by incorporating diverse data types, including clinical, socioeconomic, and psychometric variables. For example, Suh et al. developed ML models to predict early adjacent segment disease following lumbar fusion, demonstrating the potential of these tools to identify nuanced risk profiles unique to spine surgery [11]. Similarly, studies by Ogink et al. and Pedersen et al. have shown the effectiveness of ML-based models in predicting long-term recovery and complications in spine surgery patients [12, 13]. The integration of socioeconomic factors, such as income and employment status, with psychometric assessments, such as depression and anxiety levels, has further enhanced the predictive accuracy of these models [14].

Building on these findings, our study employs advanced ML techniques, including oversampling methods like SMOTE and RandomOverSampler, to improve predictive performance and address class imbalance issues inherent in clinical datasets. These methods align with the transformative potential of ML in spine surgery outcome prediction, offering clinicians robust tools for preoperative planning and personalized patient care [1, 4–7].

Despite these advancements, the application of machine learning specifically in spine surgery remains an area ripe for further exploration and innovation. There is a growing need to develop models that can better account for non-clinical variables, such as socio-economic disparities and mental health, which significantly influence surgical outcomes. Incorporating these dimensions can bridge existing gaps in spine surgery outcome prediction, offering a more holistic view of patient recovery and satisfaction [11]. Recent research highlights the potential of machine learning models to integrate and analyze diverse data sources, including preoperative clinical data, intraoperative metrics, and postoperative recovery trajectories, to enhance the precision of outcome predictions. Additionally, the inclusion of socio-economic factors such as income, job security, and healthcare access, along with psychometric assessments like stress and depression levels, provides a more robust framework for predicting both short- and long-term outcomes in spine surgery [12, 13]. These models can capture intricate patterns and relationships within the data that are often overlooked by traditional methods, thereby providing a more holistic and accurate prediction framework. Spine surgery presents unique challenges due to the intricate anatomy of the spine and the variability in surgical techniques. Additionally, the subjective nature of pain and functional recovery adds another layer of complexity to outcome prediction. The integration of AI in this field requires not only advanced algorithms but also a deep understanding of spine-specific factors, such as biomechanical properties, neurological involvement, and the impact of comorbidities [1]. Furthermore, the variability in patient recovery trajectories post-surgery necessitates

There remains a significant gap in the literature regarding the tailored application of advanced machine learning techniques and oversampling methods specifically for spine surgery outcome prediction. While some studies have explored the integration of clinical variables, fewer have focused on the predictive power of socio-economic factors, such as financial instability, and psychometric variables, such as chronic stress or anxiety, in determining long-term recovery success. Addressing this gap could significantly improve predictive accuracy by creating models that better reflect the multifaceted nature of patient outcomes [15]. This study aims to bridge this gap by developing and comparing various predictive models, including GaussianNB, ComplementNB, KNN, Decision Trees, and enhanced versions of KNN with RandomOverSampler and SMOTE. By incorporating these advanced techniques, we seek to improve the predictive accuracy and reliability of models used in spine surgery, ultimately aiding in better surgical planning and patient care [14, 16].

This research utilizes a dataset of 244 patients who underwent spine surgery, provided by the Complejo Asistencial Universitario de León (CAULE). The dataset includes a comprehensive array of variables encompassing pre-surgical, psychometric, socioeconomic, and analytical data. By applying various machine learning models and oversampling techniques, we aim to determine the most efficient predictive approach. The methodology involves detailed preprocessing, stratified data splitting, hyperparameter tuning, and rigorous model evaluation to ensure robust and reliable predictions. This structured approach is designed to offer a nuanced understanding of the factors influencing surgical outcomes, ultimately contributing to enhanced patient care and surgical planning [14, 17]. A key aspect of this study is the rigorous validation process employed to ensure the reliability and generalizability of the predictive models. Cross-validation techniques were applied to mitigate overfitting and to assess model performance across different subsets of the data. Additionally, the use of oversampling methods like SMOTE not only addresses class imbalance but also enhances the robustness of the models by generating synthetic data that reflects the complex relationships within the original dataset [18]. These steps are crucial in developing models that are not only accurate but also resilient to the variability inherent in clinical datasets.

This manuscript builds upon the foundational work presented in our previous study [19], where we initially explored the use of various machine learning models for predicting the success of spine surgeries. While the earlier work provided valuable insights into the potential of machine learning in this domain, the current study significantly extends these findings by incorporating advanced techniques such as oversampling methods (RandomOverSampler and SMOTE) and grid search optimization. These enhancements have led to a marked improvement in model performance, with the KNN model achieving an accuracy of up to 76%, a substantial increase over the results reported in the conference paper. Additionally, the current study offers a more comprehensive analysis by examining the impact of different variable groupings on predictive accuracy, further refining the application of AI in clinical decision-making. This progression underscores the substantial advancements achieved in this research, highlighting its potential to contribute more effectively to the field of spine surgery outcome prediction.

The paper is structured as follows: Section "Methodology" delineates the machine learning models employed, details the hyperparameter adjustments, and elucidates the oversampling techniques utilized. It also provides the rationale behind the selection of specific variable groupings. Section "Experiments and Results" offers a comprehensive description of the dataset and presents the experimental setup and findings. Finally, Section "Discussion" delves into the interpretation of the findings, their broader implications, and potential limitations. "Conclusions" summarizes the key takeaways of the research and outlines future research avenues and potential improvements.

## Methodology

The methodology used to carry out this research, from data collection to modelling results, is summarised in Fig. 1.

### Artificial intelligence techniques

In this study, we aim to identify the most impactful variables influencing the outcomes of spine surgery by employing a range of machine learning algorithms. The models utilized include Gaussian Naïve Bayes, Complement Naïve Bayes, K-nearest neighbors (KNN), and Decision Trees. Additionally, we applied oversampling techniques such as RandomOverSampler and SMOTE to address class imbalance in the dataset. Below, we provide an overview of each technique and the rationale for their selection, supported by relevant scientific references.

- **Gaussian Naïve Bayes:** Known for its simplicity and efficiency, Gaussian Naïve Bayes is particularly effective in scenarios with continuous data, where it assumes a normal distribution. This model's performance has been well-documented in various methodological applications, making it a suitable choice for our analysis [20].
- **Complement Naïve Bayes:** This variant of the Naïve Bayes algorithm is tailored to handle imbalanced datasets by modifying the calculation of probabilities to better represent the minority class. Its proven success in similar methodological settings justifies its inclusion in our study [21].
- **K-nearest Neighbors (KNN):** KNN is an instance-based learning algorithm that does not assume any specific distribution for the data. It has shown high effectiveness in capturing complex, non-linear relationships, which is crucial for accurate prediction in various applications [22].
- **Decision Trees:** Decision Trees provide an intuitive and visual representation of decision processes, making them valuable for identifying key variables and their interactions. Their ability to model non-linear relationships enhances their applicability in various decision-making processes [23].
- **Oversampling Methods (RandomOverSampler and SMOTE):** To mitigate the issue of class imbalance, which is common in many datasets, we employed RandomOverSampler to increase the representation of minority classes by simple replication. Additionally, SMOTE generates synthetic samples by interpolating between existing minority class samples, thus enhancing the diversity and robustness of the training data. These techniques have been shown to significantly improve model performance in imbalanced datasets [24].

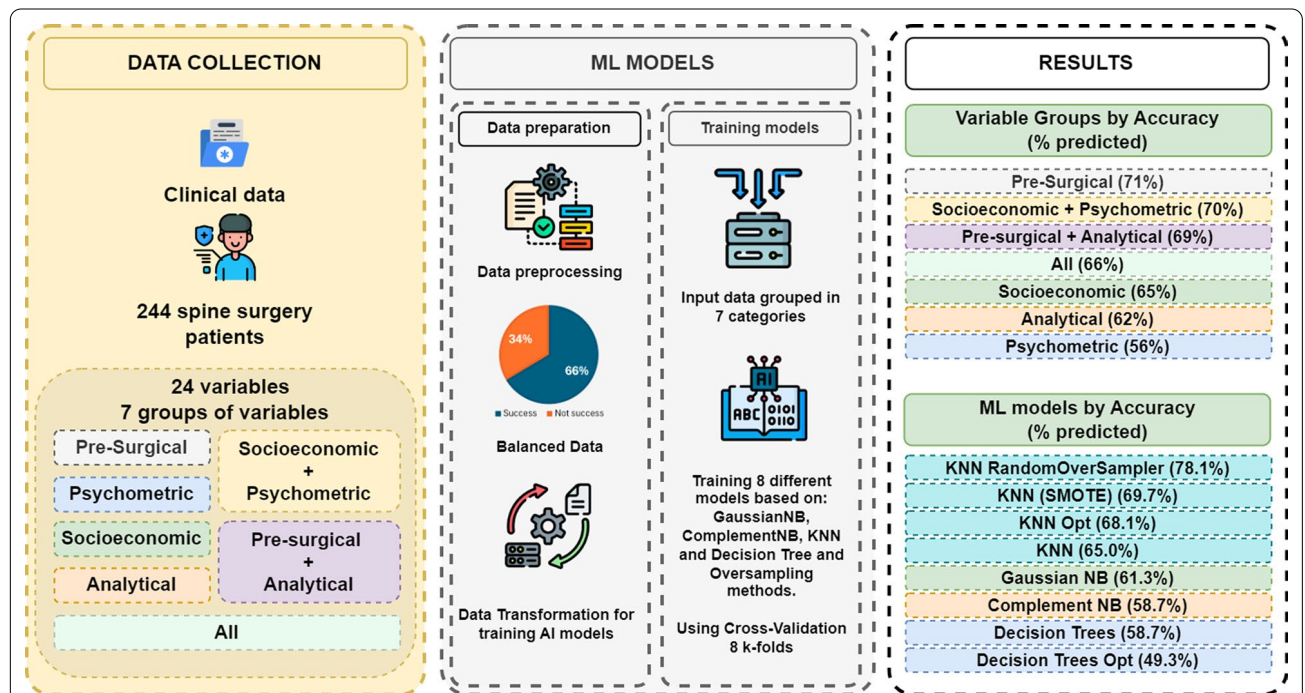

**Fig. 1** Methodology applied in research from the collection of data from spinal surgery patients to obtaining predictive models of success in spinal surgery

The selected algorithms are tailored to address the multifactorial nature of spine surgery outcomes. These techniques allow us to analyze diverse variables-socioeconomic, psychometric, and clinical-to uncover patterns and relationships that traditional methods might overlook. Each algorithm plays a specific role in predicting the success of surgeries by addressing distinct challenges, such as class imbalance, variable non-linearity, and interpretability, as detailed below.

*Gaussian Naïve Bayes*
The Gaussian Naïve Bayes classifier leverages the Bayes Theorem to compute the probability of each class based on the input features, assuming that the features follow a normal distribution. This method solves part of the problem by simplifying the prediction task. It assumes that each feature (e.g., age, psychometric score) contributes independently to the outcome, and computes the likelihood of each feature following a normal distribution. In a step-by-step manner, Gaussian Naïve Bayes first calculates the likelihood for each feature, then combines these probabilities to estimate the overall probability of each outcome (successful or unsuccessful surgery). By selecting the outcome with the highest posterior probability, this method provides an efficient, probabilistic solution for the overall prediction [25].

$$P(x) = \frac{1}{\sqrt{2\pi\sigma^2}} e^{-\frac{(x-\mu)^2}{2\sigma^2}} \qquad (1)$$

where $\mu$ is the mean:

$$\mu = \frac{1}{n} \sum_{i=1}^{n} x_i \qquad (2)$$

And $\sigma$ is the standard deviation:

$$\sigma = \sqrt{\frac{1}{n-1} \sum_{i=1}^{n} (x_i - \mu)^2} \qquad (3)$$

In this study, GNB is particularly useful for identifying patterns in analytical variables, such as pre-surgical glucose or cholesterol levels, which often follow a normal distribution. By leveraging these features, GNB provides a probabilistic baseline for understanding the relationship between metabolic health and surgical outcomes, offering insights into the broader clinical context of spine surgeries.

*Complement Naïve Bayes*
Complement Naïve Bayes modifies the traditional Naïve Bayes approach to better handle imbalanced datasets by focusing on the complement of each class. This method addresses the problem of imbalanced data, where successful outcomes may dominate the dataset. By focusing on the complement of each class, this algorithm corrects for the over-representation of the majority class and improves the prediction of minority outcomes, such as less common poor recovery cases. The step-by-step process adjusts the probability estimates to better represent these underrepresented cases, thereby providing a more balanced prediction [26].

CNB addresses the challenge of class imbalance by recalibrating probabilities to enhance the prediction of less frequent outcomes, such as unsuccessful recoveries. This is particularly relevant for psychometric variables, where depressive or somatic symptoms (e.g., MSPQ or Zung scores) might otherwise be underrepresented, enabling the model to better capture the impact of mental health on post-surgical success.

*K-nearest neighbors (KNN)*
K-nearest Neighbors (KNN) is a non-parametric method used for classification by finding the majority class among the k-nearest neighbors of a given data point. This method addresses part of the prediction problem by leveraging the similarity between patients. For each patient, KNN identifies other patients with similar characteristics (such as socio-economic and psychometric data) and predicts the outcome based on the outcomes of these similar cases. The process involves calculating the distance between the target patient and all others in the dataset, selecting the K closest neighbors, and determining the most frequent outcome (successful or unsuccessful surgery) among these neighbors. In this step-by-step manner, KNN provides a local solution, where predictions are based on the outcomes of the most similar patients [27].

KNN captures complex, non-linear relationships between pre-surgical variables (e.g., ODI scores, leg pain severity) and surgical outcomes. By comparing a patient to their closest "neighbors" in the dataset, KNN predicts outcomes based on historical patterns. For example, a patient with similar ODI and MSPQ scores to others with poor recovery can be flagged as high-risk, guiding clinicians toward targeted interventions.

*Decision trees*
Decision Trees split the dataset into subsets based on the value of input features, creating a tree-like model of decisions. This method solves part of the problem by breaking down the decision-making process into a series of smaller, manageable decisions. Each internal node in the tree represents a decision based on a feature (e.g., age, employment status), and each branch represents the possible outcomes. At each step, the algorithm selects the feature that best splits the data into groups, based on metrics like information gain or Gini index. By following the tree down to a leaf node, the model assigns the final prediction. This approach makes Decision Trees highly interpretable and suitable for identifying key variables in predicting surgical outcomes [28].

In the context of spine surgery, decision trees offer interpretability by visualizing the decision-making process. For instance, the model can identify key variables like a combination of high pre-surgical pain scores and low socioeconomic support as predictors of poor outcomes. This transparent structure provides actionable insights for healthcare professionals, facilitating personalized treatment plans.

*Oversampling methods (RandomOverSampler and SMOTE)*
RandomOverSampler and SMOTE are oversampling techniques used to address class imbalance by either replicating instances of the minority class (RandomOverSampler) or creating synthetic samples (SMOTE). While these oversampling techniques were applied specifically to KNN in this study, they were not used with GaussianNB, ComplementNB, or Decision Trees. This decision was based on the characteristics of these models: GaussianNB and ComplementNB calculate probabilities independently of class frequencies, and Decision Trees adaptively split data based on decision rules, making them less sensitive to imbalanced datasets. However, these methods effectively handle the imbalance problem by ensuring that the minority class (such as unsuccessful surgery outcomes) is adequately represented. RandomOverSampler increases the number of instances of the minority class by duplicating existing data points, while SMOTE generates new synthetic instances by interpolating between existing points. This step-by-step process ensures that the machine learning models receive a more balanced dataset, improving their ability to generalize to rare cases, which are critical in predicting poor outcomes in surgery. Future research could explore the application of oversampling techniques to these models to evaluate their impact on predictive performance [29, 30]

Oversampling methods, particularly RandomOverSampler and SMOTE, play a pivotal role in addressing the imbalance inherent in the dataset, where successful surgeries are more frequent. By ensuring that minority outcomes (e.g., unsuccessful recoveries) are adequately represented, these techniques allow models like KNN to better generalize and improve sensitivity to patients at higher risk of poor outcomes, ultimately aiding in pre-surgical decision-making.

## Experiments and results
### Dataset
The dataset was derived from a comprehensive retrospective review of clinical data from patients who underwent spine surgery and were followed up for at least six months postoperatively. A total of 244 patients were included in this study. These patients were evaluated by a multidisciplinary team consisting of orthopedic surgeons and psychiatrists. The clinical and analytical variables were meticulously collected following the approved protocol by the Complejo Asistencial Universitario de León (CAULE).

The inclusion criteria for this study were as follows:

1. Patients who underwent spine surgery for various spinal conditions.
2. Completion of at least six months of postoperative follow-up.
3. Availability of comprehensive preoperative, intraoperative, and postoperative data.

The success of the surgical outcome was defined based on patient-reported outcomes and clinical evaluations, specifically considering improvements in pain, functional status, and overall satisfaction.

The variables collected from each patient are categorized into three distinct groups:

- **Socioeconomic variables:** This group includes gender, age, and employment status.
- **Psychometric variables:** This group encompasses variables such as the presence of personal or family psychiatric history, and scores from various psychometric scales, including:
  - The MSPQ (Modified Somatic Perception Questionnaire).
  - The Zung Self-Rating Depression Scale.
  - The DRAM (Distress and Risk Assessment Method).

- **Analytical variables:** These are preoperative blood test results, including:
  - Glucose
  - Urea
  - Uric Acid
  - Creatinine
  - Cholesterol

  These parameters were selected for their routine availability in clinical practice and their known relevance to a patient's metabolic and systemic health, which can significantly influence surgical outcomes and recovery trajectories.

Preoperative variables also include specific assessments related to spine health and pain:

- Pre-surgical lumbar evaluation (Visual Analogue Scale).
- Pre-surgical leg evaluation (Visual Analogue Scale).
- Pre-surgical Oswestry Disability Index (ODI).

Postoperative variables collected six months after surgery include:

- Post-surgical lumbar evaluation (Visual Analogue Scale).
- Post-surgical leg evaluation (Visual Analogue Scale).
- Post-surgical Oswestry Disability Index (ODI).

The dataset, thus, provides a comprehensive set of variables that span clinical, psychometric, socioeconomic, and analytical domains, offering a rich basis for applying advanced machine learning techniques to predict surgical outcomes. All the variables are showed in table 1.

The dataset, carefully compiled by the Traumatology Service of the Complejo Asistencial Universitario de León, comprised exclusively numerical values, eliminating the need for any additional variable encoding to facilitate numerical processing. The success of spine surgery was determined based on the patient's outcome, assessed through two key questions: 'How satisfied were you with the surgical procedure?' and 'Would you undergo surgery again at this time?'. Each question provided five possible responses, coded as follows: 0 for 'Sure', 1 for 'Quite sure', 2 for 'I don't know', 3 for 'Quite ', and 4 for 'No'. Responses of 'Very satisfied' or 'Quite satisfied' were categorized as 'satisfactory' and assigned a code of 1, while all other responses were categorized as 'not satisfactory' and coded as 0. A successful surgical outcome, defined as 0 (Satisfactory Operation, 52.2%), was recorded if both responses were 1 or lower; otherwise, it was marked as 1 (Unsatisfactory Operation, 47.8%).

**Table 1** Description and values of variables for spine surgery dataset

| Variable | Description and values |
| --- | --- |
| Gender (GEN) | 0=Female, 1=Male |
| Age (AGE) | Numerical |
| Body Mass Index (BMI) | Numerical |
| Levels (LEVELS) | Number of instrumented levels (numerical) |
| Employment Status (EMP_ST) | 1=Regular paid work, 2=Irregular work, partial labor performance, 3=Temporary work disability, 4=Unpaid housework, 5=Unpaid housework with partial performance, 6=Student, 7=Retired, 8=Unemployment, 9=Non-disabling sequelae, 10=Partial permanent disability, 11=Total permanent disability, 12=Absolute permanent disability, 13=Permanent disability (severe disability) |
| MSPQ | Quantitative |
| ZUNG | Quantitative |
| DRAM | 0 = Group A: Normal Zung <36, 1 = Group B: Patient at risk Zung 36-51 and MSPQ <12, 2 = Group C: Affected: Depressive Zung >52, 3 = Group C: Affected: Somatic Zung 36-51 and MSPQ >12 |
| Pre-surgical Lumbar Evaluation (PRE_LUMBAR_EVA) | MIN: 0 MAX:10 (Visual Analogue Scale) |
| Pre-surgical Leg Evaluation (PRE_LEG_EVA) | MIN: 0 MAX:10 (Visual Analogue Scale) |
| 6-Month Post-surgical Lumbar Evaluation (6 M_LUMBAR_EVA) | MIN: 0 MAX:10 (Visual Analogue Scale) |
| 6-Month Post-surgical Leg Evaluation (6 M_LEG_EVA) | MIN: 0 MAX:10 (Visual Analogue Scale) |
| Pre-surgical ODI (PRE_ODI) | MIN: 0 MAX:100 (Oswestry disability test) |
| 6-Month Post-surgical ODI (6 M_POST_ODI) | MIN: 0 MAX:100 (Oswestry disability test) |
| How satisfied were you with the surgical procedure? (SAT_SURGICAL_PROC) | 0=Sure, 1=Quite sure, 2=I don't know, 3=Quite sure not, 4=No |
| Satisfaction with pain treatments (SAT_PAIN_PRE) | 0=Sure, 1=Quite sure, 2=I don't know, 3=Quit sure note, 4=No |
| 6-Month (SAT_SURGICAL_6m) | 0=Sure, 1=Quite sure, 2=I don't know, 3=Quite sure not, 4=No |
| 6-Month (SAT_PAIN_PRE)) | 0=Sure, 1=Quite sure, 2=I don't know, 3=Quite sure not, 4=No |
| Success | If COMI Question 6 and 7 at 6 M $\leq$ 1 then Success, 0=No success, 1=Success |
| Glucose (GLU) (pre-surgical) | (70-110) mg/dL |
| Urea (UREA) (pre-surgical) | (16-49) mg/dL |
| Uric Acid (URIC_ACID) (pre-surgical) | (2.4—5.7) mg/dL |
| Creatinine (CREAT) (pre-surgical) | (0.5—0.9) mg/dL |
| Cholesterol (CHOL) (pre-surgical) | (200-250) mg/dL |

**Experimental setup**

During the initial phase, data preprocessing was carried out to ensure the dataset was clean and properly formatted for analysis. The input variables were categorized into three groups as previously described: socio-economic, psychometric, and analytical. The outcome variable, referred to as "success", represents the effectiveness of the spine surgery, coded as 1 for a successful outcome and 0 for an unsuccessful one. The dataset comprised 244 patients, with a gender distribution of 47.5% female and 52.5% male. Of these patients, 52.2% experienced a successful surgical outcome, while 47.8% did not.

To prepare the data for modeling, categorical variables were encoded using a LabelEncoder, and continuous variables were standardized. Following this, the most relevant features were selected through a combination of SelectKBest and ExtraTrees algorithms, ensuring that the models were built on the most informative data.

The dataset was split into training (75%) and testing (25%) subsets using the StratifiedShuffleSplit method, which ensures that the class distribution is preserved in both subsets. The training subset was used for model training and 8-fold cross-validation, while the testing subset remained unseen during training to provide an independent evaluation of model performance.

For the experiments, the input variables were segmented into seven distinct groups shown in Table 2:

**Table 2** Division of input variables into groups

| Group | Description |
| --- | --- |
| Group I | Pre-surgical variables |
| Group II | Socioeconomic variables |
| Group III | Psychometric variables |
| Group IV | Analytical variables |
| Group V | Combination of pre-surgical and analytical variables |
| Group VI | Combination of socioeconomic and psychometric variables |
| Group VII | All variables except postoperative |

**Table 3** Nomenclature and description of models utilized in the experiments

| Model | Description |
| --- | --- |
| GaussianNB | Gaussian Naïve Bayes |
| ComplementNB | Complement Naïve Bayes |
| KNN | k-Nearest Neighbors |
| KNN_opt | KNN with optimized hyperparameters |
| KNN_RO | KNN with optimized hyperparameters and RandomOverSampler |
| KNN_SMOTE | KNN with optimized hyperparameters and SMOTE |
| DT | Decision Tree |
| DT_opt | Decision Tree with optimized hyperparameters |

The experiments consisted of applying the following models to predict the outcome variable (surgical success). The nomenclature of the models used is detailed in Table 3.

To ensure robustness and reduce the impact of randomness, an 8-fold cross-validation was employed across all experiments. Each model underwent evaluation using the most appropriate hyperparameter tuning methods tailored to the specific input data. Additionally, in certain scenarios, the oversampling techniques RandomOverSampler and SMOTE were utilized to enhance model performance. The results presented below correspond to the metrics (Accuracy and F1-Score) obtained from the independent testing subset. These metrics were calculated after training and cross-validation on the training subset, ensuring an unbiased evaluation of the models on unseen data.

**Results**

Figure 2a presents the average results for both Accuracy (Acc) and F1-Score (F1) across the seven groups of variables used in this study. These groups include pre-surgical, socio-economic, psychometric, and combinations of these, as well as the group containing all variables except post-operative data. The comparison highlights how different variable groups influence the performance of the machine learning models, showing that some combinations, such as the 'Pre-surgical + Analytical' group, yield significantly better results than others, particularly in terms of Accuracy.

Figure 2b complements this by presenting the average performance of the eight machine learning algorithms applied to all variable groups. The results indicate that some models, like K-nearest Neighbors with oversampling techniques (KNN_RO and KNN_SMOTE), consistently outperform others in terms of both Accuracy and F1-Score, especially when handling the imbalanced dataset.

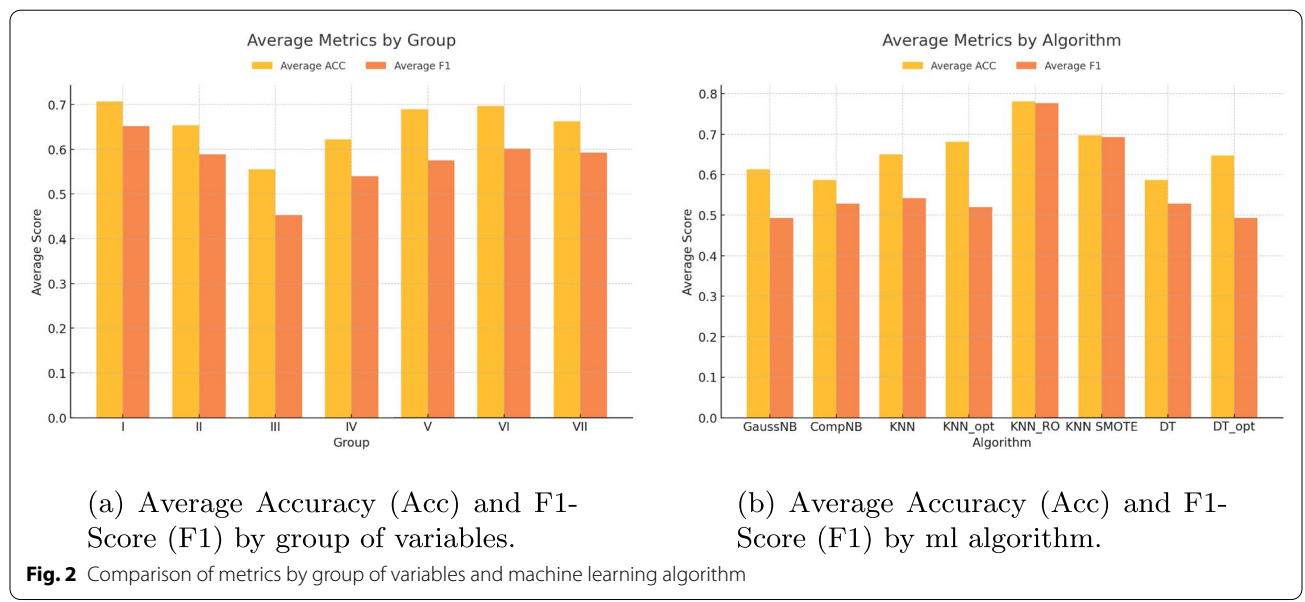

(a) Average Accuracy (Acc) and F1-Score (F1) by group of variables.

(b) Average Accuracy (Acc) and F1-Score (F1) by ml algorithm.

**Fig. 2** Comparison of metrics by group of variables and machine learning algorithm

These figures not only summarize the numerical performance of the models but also represent the core concept of our work: the integration of advanced machine learning techniques, such as KNN with oversampling (KNN_RO), and the strategic combination of pre-surgical and analytical variables to achieve superior predictive accuracy and F1-Score. The results in Fig. 2b underline the transformative potential of addressing dataset imbalances through oversampling methods, which enhance the model's sensitivity to less frequent outcomes (e.g., unsuccessful recoveries).

Furthermore, the comparison in Fig. 2a highlights the importance of selecting specific variable combinations, such as the pre-surgical and analytical variables group, which yielded the best accuracy metrics. This emphasizes how our approach advances beyond traditional models by leveraging non-clinical variables and novel preprocessing techniques to improve spine surgery outcome predictions.

This conceptual representation demonstrates the broader implications of the methodology, providing a pathway for implementing these techniques in clinical decision-making processes.

Additionally, Table 4 provides the detailed Accuracy (Acc) and F1-Score (F1) metrics obtained for each model across all experiments.

The analysis reveals some noteworthy findings. For instance, the KNN_RO model achieves the highest Accuracy of 88% in group V (combination of pre-surgical and analytical variables), while the KNN model reaches the highest F1-Score of 67% in group I (Pre-surgical variables). This superior performance of KNN_RO can be attributed to its ability to handle imbalanced data through the RandomOverSampler technique, which replicates minority class instances, effectively improving the model's sensitivity to rare outcomes (e.g., poor surgery recovery). Moreover, the use of cross-validation techniques (8-fold) ensured that the results are robust and not biased by a particular partition of the dataset, further supporting the reliability of the model's performance. In comparison, models like Gaussian Naïve Bayes underperform due to their assumption of feature independence, which limits their capacity to capture complex relationships between variables such as psychometric and socio-economic factors.

When grouping the F1-Score metrics obtained for each group of variables across all experiments, the means and standard deviations are displayed in Table 5. The group of variables with the highest mean F1-Score is the combination of socio-economic and psychometric variables, with a mean F1-Score of 0.66. This finding suggests that these combined variables provide a strong predictive signal when integrated into machine learning models, likely due to the interplay between mental health indicators and financial or social status, which have a direct impact on recovery outcomes. Moreover, the models' ability to capture these relationships-especially those like KNN with oversampling-demonstrates the importance of selecting and combining variables effectively to improve predictive accuracy.

**Table 4**  Accuracy (Acc) and F1-Score (F1) metrics obtained by applying the 8 machine learning models to the 7 groups of variables

| Model | I | II | III | IV | V | VI | VII |
|---|---|---|---|---|---|---|---|
| GaussNB (Acc) | **0.63** | 0.63 | 0.65 | 0.53 | 0.63 | 0.63 | 0.59 |
| GaussNB (F1) | 0.55 | 0.47 | 0.49 | 0.46 | 0.47 | **0.53** | 0.48 |
| CompNB (Acc) | 0.59 | 0.59 | 0.45 | 0.57 | 0.63 | **0.71** | 0.57 |
| CompNB (F1) | 0.55 | 0.57 | 0.43 | 0.55 | 0.47 | **0.61** | 0.52 |
| KNN (Acc) | **0.76** | 0.63 | 0.53 | 0.59 | 0.65 | 0.73 | 0.66 |
| KNN (F1) | **0.67** | 0.58 | 0.38 | 0.50 | 0.49 | 0.61 | 0.56 |
| KNN_opt (Acc) | **0.78** | 0.69 | 0.61 | 0.63 | 0.71 | 0.67 | 0.68 |
| KNN_opt (F1) | **0.69** | 0.62 | 0.38 | 0.47 | 0.56 | 0.40 | 0.52 |
| KNN_RO (Acc) | 0.82 | 0.82 | 0.56 | 0.77 | **0.88** | 0.83 | 0.79 |
| KNN_RO (F1) | 0.81 | 0.82 | 0.56 | 0.76 | **0.88** | 0.83 | 0.78 |
| KNN SMOTE (Acc) | **0.80** | 0.67 | 0.56 | 0.65 | 0.74 | 0.67 | 0.79 |
| KNN SMOTE (F1) | **0.80** | 0.66 | 0.56 | 0.63 | 0.74 | 0.67 | 0.79 |
| DT (Acc) | 0.59 | 0.59 | 0.45 | 0.57 | 0.63 | **0.71** | 0.57 |
| DT (F1) | 0.55 | 0.57 | 0.43 | 0.55 | 0.47 | **0.61** | 0.52 |
| DT_opt (Acc) | **0.60** | 0.42 | 0.39 | 0.40 | 0.52 | 0.55 | 0.57 |
| DT_opt (F1) | 0.60 | 0.61 | 0.63 | **0.67** | 0.65 | 0.63 | 0.65 |

The best values for each metric (Acc and F1) for each model applied are highlighted in bold

## Discussion

The results of our experiments underscore the significance of combining different variable types-particularly socioeconomic and psychometric variables-in enhancing the predictive performance of machine learning models for spine surgery outcomes. Compared to [19], which focused primarily on traditional clinical variables, our study integrates socio-economic and psychometric data to provide a more comprehensive predictive framework. This distinction highlights the added value of non-clinical variables in capturing the multifactorial nature of patient recovery and improving predictive accuracy. Moreover, our inclusion of oversampling techniques like RandomOverSampler and SMOTE further differentiates this work, as these methods were not explored in [19], emphasizing the importance of addressing class imbalance in medical datasets. The findings align with existing literature, which suggests that non-clinical factors can play a crucial role in predicting surgical success. Specifically, our study revealed that the combination of socioeconomic and psychometric variables yielded the highest mean F1-Score of 0.66, suggesting that these variables provide complementary information that is critical for accurate prediction. However, beyond the influence of these variables, the application of advanced machine learning techniques, such as K-nearest Neighbors with oversampling (KNN_RO), played a pivotal role in achieving higher predictive accuracy. These techniques addressed the imbalance in the dataset, allowing the models to generalize better across both majority and minority classes, thus enhancing their overall robustness and reliability.

Our results further demonstrate that integrating advanced machine learning (ML) techniques and oversampling methods, such as SMOTE and RandomOverSampler, leads to substantial performance improvements over previous studies. For instance, the KNN model with RandomOverSampler in our study achieved an accuracy of up to 88% and an F1-score of 0.81, outperforming the accuracy of 82% reported by Greenberg et al., who utilized synthetic data derivatives for spine surgery predictions [6]. Similarly, compared to Tragaris et al., who achieved moderate improvements using traditional ML approaches in spine surgery [7], our models leveraged the integration of diverse variable types and oversampling techniques to capture complex, multifactorial relationships in patient outcomes.

**Table 5**  Mean Accuracy (Acc) and F1-Score (F1) with standard deviation (SD) obtained according to the group of variables used in all experiments

| Variable group | Mean Acc | SD Acc | Mean F1 | SD F1 |
|---|---|---|---|---|
| Pre-surgical | **0.71** | 0.09 | 0.65 | 0.10 |
| Socioeconomic | 0.65 | 0.07 | 0.59 | 0.11 |
| Psychometric | 0.56 | 0.07 | 0.45 | 0.07 |
| Analytical | 0.62 | 0.07 | 0.54 | 0.11 |
| Pre-surgical + Analytical | 0.69 | 0.08 | 0.58 | 0.14 |
| Socioeconomic + Psychometric | 0.70 | 0.07 | **0.66** | 0.14 |
| All except postoperative | 0.66 | 0.09 | 0.59 | 0.13 |

Bold highlights the best Mean ACC and best Mean F1

These improvements are consistent with previous findings in the literature, such as Suh et al., who emphasized the value of incorporating socioeconomic and psychometric data for nuanced predictions in adjacent segment disease following lumbar fusion [11]. By integrating psychometric assessments, such as depression and anxiety scores, alongside preoperative clinical variables, our study achieved a mean F1-score of 0.66 for combined socioeconomic and psychometric variables, exceeding the 62% F1-score reported by Ogink et al. using similar ML models [12].

The KNN_RO model, which applied RandomOverSampler to address class imbalance, demonstrated superior performance across several groups of variables. While oversampling techniques were applied exclusively to KNN, this decision reflects the sensitivity of KNN to class imbalance due to its reliance on local neighborhoods for decision-making. However, future work could explore the application of RandomOverSampler and SMOTE to GaussianNB, ComplementNB, and Decision Trees to determine if similar improvements can be achieved with these models. This highlights the effectiveness of addressing class imbalance, particularly in medical datasets where positive outcomes may be less frequent. The ability of KNN_RO to adapt to the dataset's characteristics through oversampling and its capacity to model non-linear relationships between variables (e.g., pre-surgical and analytical factors) underscores the importance of selecting both appropriate models and preprocessing techniques. These techniques, when optimized, significantly improve the model's sensitivity to rare outcomes and enhance its ability to capture complex patterns in the data. The enhanced performance of KNN_RO, particularly in Group V (a combination of pre-surgical and analytical variables), highlights how oversampling techniques like RandomOverSampler can improve model robustness by ensuring that the minority class (successful outcomes) is adequately represented.

Interestingly, while the analytical variables group achieved the highest F1-Score in the DT_opt model (67%), its overall performance was less consistent across other models, as reflected by its lower average F1-Score of 54%. This suggests that while Decision Trees with optimized hyperparameters can perform well in specific cases, their reliance on precise decision rules makes them less flexible in capturing non-linear relationships, especially when the dataset includes diverse variable types. In contrast, models like KNN, which do not assume a strict structure in the data, exhibit greater adaptability and robustness across different groups of variables, particularly when paired with effective preprocessing techniques. The socioeconomic and psychometric variables, when combined, exhibited more consistent performance, further highlighting their robustness in predicting surgical outcomes.

The high performance of the KNN_RO model, especially in Group V (pre-surgical and analytical variables), underscores the potential of advanced machine learning techniques in improving predictive accuracy in spine surgery. This result reinforces the importance of integrating non-linear models that can leverage both patient-specific and broader analytical data, offering a more holistic perspective on predictive modeling. The success of KNN_RO also emphasizes the need for models that can handle real-world medical data complexities, such as class imbalance, through techniques like oversampling. This result is particularly relevant as it suggests that integrating patient-specific pre-surgical data with broader analytical data can provide a more holistic view, leading to better predictions.

Moreover, the success of the KNN model in achieving the highest F1-Score of 67% within the pre-surgical variables group suggests that patient-specific factors assessed before surgery are crucial indicators of surgical outcomes. This aligns with previous studies that have emphasized the importance of preoperative assessments in determining the likelihood of surgical success.

The variability in performance across different groups of variables also highlights the multifactorial nature of spine surgery outcomes. While traditional clinical assessments remain valuable, the integration of socioeconomic and psychometric factors provides a more comprehensive approach, reflecting the complex interplay of various factors influencing patient recovery and overall satisfaction. This further demonstrates the importance of model flexibility and the ability to adapt to complex, high-dimensional datasets through techniques like KNN and Decision Trees, which can capture both linear and non-linear relationships between variables.

These findings are consistent with the broader literature on the use of machine learning in clinical outcome prediction. In particular, this study advances the methodology presented in [19] by incorporating socio-economic and psychometric variables alongside clinical data and leveraging oversampling techniques such as RandomOverSampler and SMOTE to improve predictive performance. This approach underscores the importance of methodological innovations in enhancing the robustness and generalizability of machine learning models in clinical settings. For instance, previous studies have shown that incorporating diverse data types, including non-clinical factors, can significantly enhance the predictive accuracy of models in various surgical fields [31, 32]. Our results contribute to this body of work by demonstrating the specific utility of these approaches in spine surgery. Moreover, this study reinforces the importance of selecting the appropriate machine learning techniques and preprocessing strategies, such as oversampling and cross-validation, to ensure that models are robust and generalizable in medical contexts.

In conclusion, our study highlights the importance of considering a wide range of variables, including non-clinical factors, in the prediction of spine surgery outcomes. The use of advanced machine learning techniques, coupled with effective data preprocessing and oversampling methods, has the potential to significantly improve the accuracy and reliability of predictive models, ultimately leading to better surgical planning and improved patient care. Future research should continue to explore the integration of diverse data sources and the development of adaptive models that can account for the dynamic nature of patient recovery trajectories. Additionally, future investigations could focus on applying oversampling techniques, such as RandomOverSampler and SMOTE, to models like GaussianNB, ComplementNB, and Decision Trees. This would provide a more comprehensive understanding of how these methods influence model performance in imbalanced datasets and extend the applicability of the findings in this study.

## Conclusions

This study offers significant insights into the application of advanced machine learning techniques for predicting outcomes in spine surgery. A key innovation of our research is the use of variables that can be determined preoperatively, including socioeconomic status and psychometric factors, which allows for the prediction of surgical outcomes before the procedure takes place. This pre-surgical predictive capability is particularly valuable, as it enables clinicians to make more informed decisions regarding surgical planning and patient management, potentially improving overall outcomes.

Our findings highlight the critical role of integrating a diverse range of these preoperative variables, demonstrating that models incorporating non-clinical factors-such as socioeconomic and psychometric data-can significantly enhance predictive accuracy. The superior performance of these models suggests that a more holistic approach to pre-surgical assessment, one that considers a broader spectrum of patient information, is essential for optimizing surgical planning and improving patient outcomes.

However, several limitations must be acknowledged. The retrospective nature of our study and the constraints inherent in the dataset used present challenges. From an initial pool of 300 patients, only 244 met the inclusion criteria and had comprehensive follow-up data available. This sample size, while sufficient for preliminary analysis, may not fully represent the broader population undergoing spine surgery. Additionally, the dataset, sourced from a non-randomized clinical review, may introduce biases that could be mitigated in a prospective or randomized study design. Another limitation is the exclusion of patients with certain comorbidities or severe psychiatric conditions, which may have skewed the psychometric data analyzed.

Despite these limitations, our study underscores the potential of machine learning to advance predictive analytics in spine surgery, particularly through the use of preoperatively determinable variables. As demonstrated in other fields, smaller datasets, when handled with rigorous validation techniques, can still yield meaningful and robust conclusions. Previous studies, such as those by [33–36], have shown that detailed, well-curated datasets can provide valuable insights, even in specialized medical contexts. Our work aligns with this precedent, suggesting that innovative methodological approaches can overcome the challenges posed by limited data availability.

In conclusion, while further research is necessary to validate and expand upon these findings, this study contributes to the growing body of evidence supporting the use of machine learning in clinical decision-making. The novelty of our approach lies in the integration of preoperatively available variables, which can offer significant advantages in planning and tailoring patient-specific interventions. Future studies should aim to incorporate larger and more diverse patient populations and continue to refine the integration of clinical, socioeconomic, and psychometric variables. Such efforts will be crucial in developing more accurate, personalized predictions for patients undergoing spine surgery, ultimately enhancing patient care and surgical outcomes.

**Author contributions**
José Alberto Benítez-Andrades: Conceptualization, Methodology, Software, Visualization, Validation, Writing- Original draft preparation. Camino Prada-García: Conceptualization, Supervision, Writing- Reviewing and Editing. Nicolás Ordás-Reyes: Methodology, Data curation, Software, Visualization, Writing- Reviewing and Editing. Marta Esteban Blanco: Data curation, Writing- Original draft preparation. Alicia Merayo: Conceptualization, Methodology, Writing- Reviewing and Editing. Antonio Serrano-García: Conceptualization, Data curation, Methodology, Validation, Writing- Reviewing and Editing.

**Funding**
Open Access funding provided thanks to the CRUE-CSIC agreement with Springer Nature. This work is a result of the project "NLP-Driven Insight Engine for Suicide Attempt Detection in EHR (SUICIDETECT)", that is being developed under grant "PID2023-146168OA-I00" from the Spanish Ministerio de Ciencia, Innovación y Universidades.

**Data availability**
The data that support the findings of this study are not openly available due to reasons of sensitivity and are available from the corresponding author upon reasonable request.

## Declarations

**Conflict of interest**
The authors declare that they have no competing interests.

**Author details**
[1]SALBIS Research Group, Department of Electric, Systems and Automatics Engineering, Universidad de León, Campus of Vegazana s/n, 24071 León,

Spain. ²Instituto de Investigación Biosanitaria de León (IBIOLEÓN), Calle Altos de Nava, s/n, 24008 León, Spain. ³Department of Preventive Medicine and Public Health, University of Valladolid, 47005 Valladolid, Spain. ⁴Dermatology Service, Complejo Asistencial Universitario de León, 24008 León, Spain. ⁵Department of Electric, Systems and Automatics Engineering, Universidad de León, Escuela de Ingenierías Industrial, Informática y Aeroespacial, Campus de Vegazana s/n, 24071 León, Spain. ⁶Department of Orthopaedic Surgery and Traumatology, Complejo Asistencial Universitario de León, Spain, León 24008. ⁷Psychiatry Service, Department of Psychosomatic, Complejo Asistencial Universitario de León, 24008 León, Spain.